\documentclass[10pt,twocolumn,letterpaper]{article}

\usepackage{cvpr}
\usepackage{times}
\usepackage{graphicx} % more modern
\usepackage{subfigure}
\usepackage{caption}
\usepackage{epsfig}
\usepackage{amsmath}
\usepackage{lib}
\usepackage{amssymb}
\usepackage{float}
\usepackage{epigraph}
\usepackage{pdfpages}
\usepackage{algorithm}
\usepackage{algorithmic}

\usepackage{booktabs}
\usepackage[pagebackref=true,breaklinks=true,letterpaper=true,colorlinks,bookmarks=false]{hyperref}

\cvprfinalcopy % *** Uncomment this line for the final submission

\def\cvprPaperID{672} % *** Enter the CVPR Paper ID here

\begin{document}

\title{Shape and Symmetry Induction for 3D Objects}

\author{Shubham Tulsiani$^1$, Abhishek Kar$^1$, Qixing Huang$^2$, Jo\~{a}o Carreira$^1$ and Jitendra Malik$^1$ \\
$^1$University of California, Berkeley $~~~^2$Toyota Technological Institute at Chicago\\
{\tt\small $^1$\{shubhtuls, akar, carreira, malik\}@eecs.berkeley.edu $~~~ ^2$huangqx@ttic.edu}
}

\maketitle
\begin{abstract}
Actions as simple as grasping an object or navigating around it require a rich understanding of that object's 3D shape from a given viewpoint. In this paper we repurpose powerful learning machinery, originally developed for object classification, to discover image cues relevant for recovering the 3D shape of potentially unfamiliar objects. We cast the problem as one of local prediction of surface normals and global detection of 3D reflection symmetry planes, which open the door for extrapolating occluded surfaces from visible ones. We demonstrate that our method is able to recover accurate 3D shape information for classes of objects it was not trained on, in both synthetic and real images. 
\end{abstract}

\epigraph{``What specifies an object are invariants that are themselves `formless'"}{\textit{J.J. Gibson}}

\section{Introduction}

% A. what we do
In this paper we develop a method for understanding the 3D shape of an unfamiliar object from a single image. Our method recovers a 2.5D shape representation by densely labeling normals of object surfaces visible in the image. We target the remaining 0.5D -- the shape of occluded surfaces -- by inferring shape self-similarities. As one small step in this direction, we introduce the task of detecting the orientation of any planes of reflection symmetry in the 3D object shape.

% D. how we do it
% 3. 3D shape symmetries from 2D images
Recovering 3D object shape from a single image is clearly an ill-posed problem and requires assumptions to be made about the shape. The problem of reconstructing familiar categories has seen some success, but there strong 3D priors can be learned from training data \cite{cashman2012shape,kar2015category}. The problem of reconstructing shapes for previously unseen categories is more subtle and coming up with the right priors seems critical, in particular for recovering occluded surfaces. We pursue patterns of self-similarity in 3D shape given just the image, hoping these will allow filling in occluded geometry with carefully placed copies of visible geometry. We define an entry-level version of the problem: detecting the 3D orientation of planes of reflection symmetry. %We also introduce evaluation methodology for this task and propose a multi-label ConvNet classifier that shows promising performance. 

\begin{figure}[t]
\centering
\includegraphics[width = 0.48\textwidth]{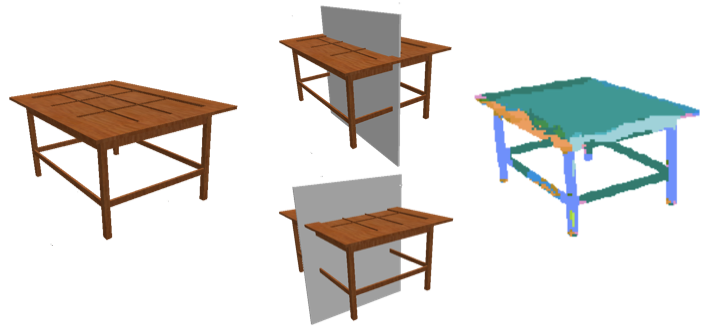}
\caption{Given a single image of a novel object, our model induces a pixel-wise labeling of its surface normals (right) and predicts the orientations of all 3D planes of reflection symmetry (center). }
\figlabel{introFig}
\end{figure}

% B. how we do it
% 1. the problem of data 
Both components of our approach rely heavily on learning large nonlinear classifiers end-to-end, which has become an effective solution to many vision problems, but not yet for 3D object shape reconstruction from a single image. This is due, in part, to technical and logistical difficulties involved in creating large datasets of images of objects with aligned 3D shapes \cite{xiang2014beyond,carreira2015lifting}.

In this work we leverage synthetic data, and resort to new large-scale shape collections where ground truth symmetry planes and surface orientations can be accurately computed. We then render these models and use the images paired with the symmetry and normal labels to learn Convolutional Neural Network (CNN) \cite{neocognitron,Lecun1989}  based systems for symmetry prediction (\secref{symmetry}) and normal estimation (\secref{normals}). We show qualitative results on real images and empirically demonstrate the ability of our models to induce these predictions accurately for novel objects (\secref{experiments}).

%We tackle a very difficult problem and there are few methods to compare with. Nevertheless, in experiments on rendered images our method demonstrates large quantitative improvement over a silhouette-based reconstruction technique \cite{twarog2012playing}, on the task of surface normal labeling. We also obtain accurate 3D symmetry plane detection for unfamiliar objects from a large range of viewpoints. Qualitative results on real images validate the practical applicability of our proposed method.

\section{Related Work}
\seclabel{related}
Recovering an object's surface geometry from a single image is an ill-posed problem in general -- e.g. a same image can be caused by different configurations of surface geometry, reflectance and lighting conditions. %Inferring 3D shape (e.g, depth/normal) and structure (e.g., symmetries) of objects depicted in a single image .
This problem has been studied from many different perspectives, which we will roughly divide here into physics-based and predictive shape inference. These two paradigms, together with prior work on symmetry detection and learning from CAD model datasets are briefly summarized below.

\paragraph{Physics-based Shape Inference}

Given an image, the preference for any particular 3D shape depends strongly on the type of priors imposed. Physics-based approaches, such as early shape from shading techniques \cite{horn1989obtaining}, aimed to optimize shape using variational formulations with regularizers that encoded strong assumptions about albedo and illumination. Modern approaches such as SIRFS \cite{BarronTPAMI2015} extend these by using richer priors and additionally reasoning over reflectances in the solution space.

 %Useful cues include lines, shading and symmetries. We refer to~\cite{palmer1999vision} for a thorough discussion of the cues that are helpful for geometry recovery, and to~\cite{conf/iccv/ThrunW05,conf/cvpr/CohenZSP12,Tao:2015:DFS} for some recent achievements of computational methods. However, the performance of these techniques heavily rely on the success of detecting and aggregating these cues, which turn out to be brittle if hand-designed as opposed to learned directly from data.

%Medial Axis Transform [???]

%Early Work \cite{cham1996geometric}

%Affine Invariant Image Symmetry \cite{cornelius2007efficient}

%Learning Method \cite{tsogkas2012learning}

%\paragraph{Learning}
%CNN \cite{neocognitron,Lecun1989} \\

%Attributes \cite{farhadi2009describing,lampert2009learning}

\paragraph{Predictive Shape Inference.} The other common paradigm for shape inference is through supervised learning techniques that leverage training data to boost inference results. Early work on predicting depth (and/or surface normals) utilized graphical models \cite{Hoiem:2005:APP,Saxena:2009:MLS} and more recent work have improved performance via hierarchical feature extractors \cite{DBLP:conf/nips/EigenPF14}. These previous approaches, however, have focused on inferring scene-level information which differs from our goal of perceiving the shape of objects.

Predicting object pose is another task for which many learning-based methods have been developed. Traditional approaches focused on particular instances and used explicit 3D models \cite{huttenlocher1990recognizing}, but the task has recently evolved  into the prediction of category-level pose \cite{vpsKpsTulsianiM15,pepik12dpm,ghodrati14viewpoint}. Category-specific pose prediction models have still the inconvenience that they do not generalize to novel object categories and require large amounts of labeled data for each of their training categories. Recently, Tulsiani \etal \cite{poseInductionTCM15} showed that treating pose as an attribute \cite{farhadi2009describing,lampert2009learning} and training a prediction system accordingly can enable prediction for unfamiliar classes. Our work aims for a similar generalization to arbitrary objects, but the symmetry and surface orientation representations we infer are much more general and detailed.

\paragraph{Learning from CAD Model Collections.} There has been a growing trend of using 3D CAD model renderings to aid computer vision algorithms. The key advantage of these approaches is that it is easy to obtain labeled training data at scale. Examples include approaches for aligning 3D models to images \cite{Aubry14, lpt2013ikea}, object detection~\cite{DBLP:journals/corr/PengSAS14} and pose estimation~\cite{renderCNN}. In this work, we apply this idea to normal estimation and symmetry detection - both easily obtained from 3D models.

%\paragraph{Symmetries in 3D}
%Survey \cite{liu2010computational}

%Symetry Detection Code \cite{mgp_approx_symm_sig_06, mgp_symmetrization_sig_07}

%Symmetric Scenes \cite{koser2011dense}
%SfM  \cite{snavely2006photo} incoportating symmetries \cite{cohen2012discovering, ceylan2014coupled}

%Single Depth Image - retrieval based symmetry {rock2015completing}

%\paragraph{Applications.}

%Car Reconstruction \cite{sinha2012detecting} \\

%Interactive Object Reconstruction \cite{Sweep3} \\

%Puffball \cite{twarog2012playing} \\ 

\section{Symmetry Prediction}
\seclabel{symmetry}

\epigraph{"Symmetry is what we see at a glance; based on the fact that there is no reason for any difference."}{- Blaise Pascal, \textit{Pens\`{e}es}}

Most real-world shapes possess symmetries. For example, all object categories available in popular datasets such as PASCAL VOC and Microsoft COCO exhibit at least bilateral reflection symmetry. Symmetry detection provides cues into the elongation modes and 3D orientation of objects which can influence perceived shape (as illustrated by Ernst Mach square/diamond famous example \cite{palmer1999vision}) and is conjectured to aid grouping \cite{koffka2013principles} and recognition \cite{vetter1997linear} in human vision. Symmetry-based approaches such as Blum's Medial Axis Transform \cite{blum67} spawned entire subcommunities \cite{siddiqi1999shock} devoted to their development.

Symmetry is however now rarely pursued in practical vision systems, perhaps because too much emphasis has been placed on``retinal" symmetries -- symmetries in planar shapes, that are only moderately distorted when projected into an image. Most objects are not planar and their symmetries can be widely deformed after projecting on to images due to the angle relative to the camera and the geometry of central projection. Consequently, we deviate from the existing techniques for detecting retinal symmetrical structures which seek dense correspondences across feature points, aiming to detect subsets of correspondences that can be realized by the underlying structures ~\cite{LiuCurveSymACCV10,Ceylan:2014:CSS,DBLP:journals/ftcgv/LiuHKG10}. Instead, we present a learning-based framework to directly detect the underlying 3D symmetries for an object.

We propose to infer representations similar to those of existing approaches that rely on 3D shape inputs \cite{mgp_symmetrization_sig_07} or depth images \cite{conf/iccv/ThrunW05, rock2015completing} - but we aim to do so from a single RGB image. In fact, we leverage existing approaches for detecting 3D symmetries in 3D meshes in order to obtain ground truth symmetries. We can then frame a supervised learning problem using rendered images of these shapes as input. We describe our symmetry extraction, symmetry prediction formulation and learning framework below.

%It is then unsurprising that symmetry has been widely studied in computer vision, tracing back all the way to seminal work by Marr \cite{}.

\paragraph{Extracting Symmetries from Shapes.}
We follow the procedure outlined by Mitra \etal \cite{mgp_symmetrization_sig_07} to extract the global reflectional symmetries given a shape. We sample the shape uniformly to correct for any biased sampling in the original mesh points. We then consider many symmetry plane hypotheses, parametrized as $(n,b)$ where the points satisfying $n \cdot x = b$ lie on the plane, and iteratively refine each hypothesis via ICP between original and reflected points. We finally discard planes that do not fit the sampled points well and additionally suppress duplicate planes with very similar orientations. We refer the reader to \cite{mgp_symmetrization_sig_07} for the exact mathematical formulation. 

\paragraph{Formulation.}
Given an image $I$, we aim to predict the symmetry planes of the underlying 3D object. Since the exact placement of a plane only assumes a meaning once we have inferred a reconstruction for the object and is not well defined given a single image, we focus on inferring the orientations of the underlying symmetry planes. Let $\mathcal{N}$ represent the space of unit norm 3D orientation vectors. We first discretize this space via approximately uniform samples on the unit sphere~\cite{swinbank2006fibonacci} $\{n_1, \cdots, n_K \}$. Our learning task is modeled as a multilabel classification where we aim to learn a mapping $f'$ s.t. $f(I) \in \{0,1\}^K $ and $f(I)[k] = 1$ iff $n_k$ is a correct discretization for some symmetry orientation of the underlying 3D object.

%Given a shape model $S$, with underlying symmetry planes $\{P_i = (n_i, b_i)\}$ as detected above

\paragraph{Learning.}

As mentioned earlier, we rely on a large shape collection to learn prediction of symmetries. We also use a rendering engine $E$ which, given a  shape model $S$ and a model rotation $R$ yields a rendered image $E(S,R)$. To obtain training data for our task, we repeatedly sample a shape $S$ belonging to some object category $c \in \mathcal{C}$ from the shape collection and detect the underlying reflectional symmetry planes $\{P_i = (n_i, b_i)\}$ as described above. We then sample a model pose from a view distribution $\mathcal{V}$ and obtain the rendered image $E(S,R)$. The symmetry orientations underlying the 3D shape of the rendered image ($\{R \ast n_i\}$) are computed by rotating the symmetry orientations of the shape $S$. These orientations are discretized as into orientation bins described above to obtain a label $l \in  \{0,1\}^K$. The pair $(E(S,R), l)$ forms one training exemplar for our problem. We sample models and views repeatedly to generate the training data - the exact details are described in the experiments.

Given the training set constructed above, we train a CNN to predict symmetries given a single image. More concretely, we use an Alexnet \cite{krizhevsky2012imagenet} based architecture with $K$ outputs in the last layer and use a sigmoid cross entropy loss to enforce the outputs to represent log-probability of the corresponding orientation being a symmetry plane for the underlying 3D object. Note that the system is trained in a category-agnostic way \i.e. unlike common detection and pose prediction systems \cite{vpsKpsTulsianiM15, renderCNN} , we share output units across all object categories $c \in \mathcal{C}$. This implicitly enforces the CNN based symmetry prediction system to exploit similarities across object classes and learn common representations that may be useful for generalizing to novel objects. Our experiments empirically demonstrate that the system we describe is indeed capable of predicting symmetries for objects belonging to a category $c \notin \mathcal{C}$.

\section{Surface Normal Estimation}
\seclabel{normals}

The importance of \textit{perceiving the surface layout} was highlighted by Gibson as early as 1950 \cite{gibson1950perception}.  These ideas were grounded more computationally as Marr's 2.5D sketch representations \cite{marr1982vision}. Koenderink, Van Doorn and Kappers later demonstrated \cite{koenderink1996pictorial}  the ability of humans to recover surface orientations from pictures and shaded objects. All these seminal works, perceptual as well as computational, emphasized the importance of perceiving surface orientations as an integral part of perception.

Single-image depth \cite{DBLP:conf/nips/EigenPF14,Saxena:2009:MLS,Karsch:TPAMI:14} and surface normal \cite{fouheyICCV13,eigeniccv15,Wang15} prediction using CNNs has shown promise when dealing with the shape of scenes. Scenes exhibit strong regularities: the ground and the ceiling is horizontal, the walls are vertical. Here we demonstrate that these models can be leveraged to label the much more complex normals of object surfaces. We describe our formulation and learning procedure below.

\paragraph{Formulation.}
Our aim is to learn a model that is capable of constructing a mapping from pixels to orientations given an image $I(\cdot,\cdot)$. The desired output, given the input image $I$ is a spatial orientation function $N(\cdot,\cdot)$ such that $N(x,y)$ is the surface orientation of the point in the underlying 3D shape that is projected at pixel $(x,y)$ in the given image.
Instead of directly predicting an orientation $n \in \mathcal{N}$ at each spatial location, we follow a formulation motivated by Koenderink's experiment where the subjects were able to reconstruct a dense sampling of surface orientations in images using an element from discrete set of \textit{gauge figures} placed at every location. This discretization of surface orientations has been previously successfully leveraged \cite{zeisl2014discriminatively} for estimating surface normals of a scene. Our intuition is that this approach combined with CNN architectures that have shown rapid recent progress for pixelwise classification tasks \eg semantic segmentation can yield promising results in the domain of object shape perception.

Operationally, similar to our approach for symmetry plane orientations, we discretize the space of visible surface orientations into $K$ discrete bins using approximately uniform samples over the half unit sphere \cite{swinbank2006fibonacci}. The goal for normal estimation is to then learn a function approximation $f$ s.t $f(I) = N$ where $N(x,y)$ assigns the correct orientation bin for the 3D point projected at $(x,y)$.

\seclabel{experiments}
\begin{figure*}[t!]
  \centering

\includegraphics[width=0.95\textwidth]{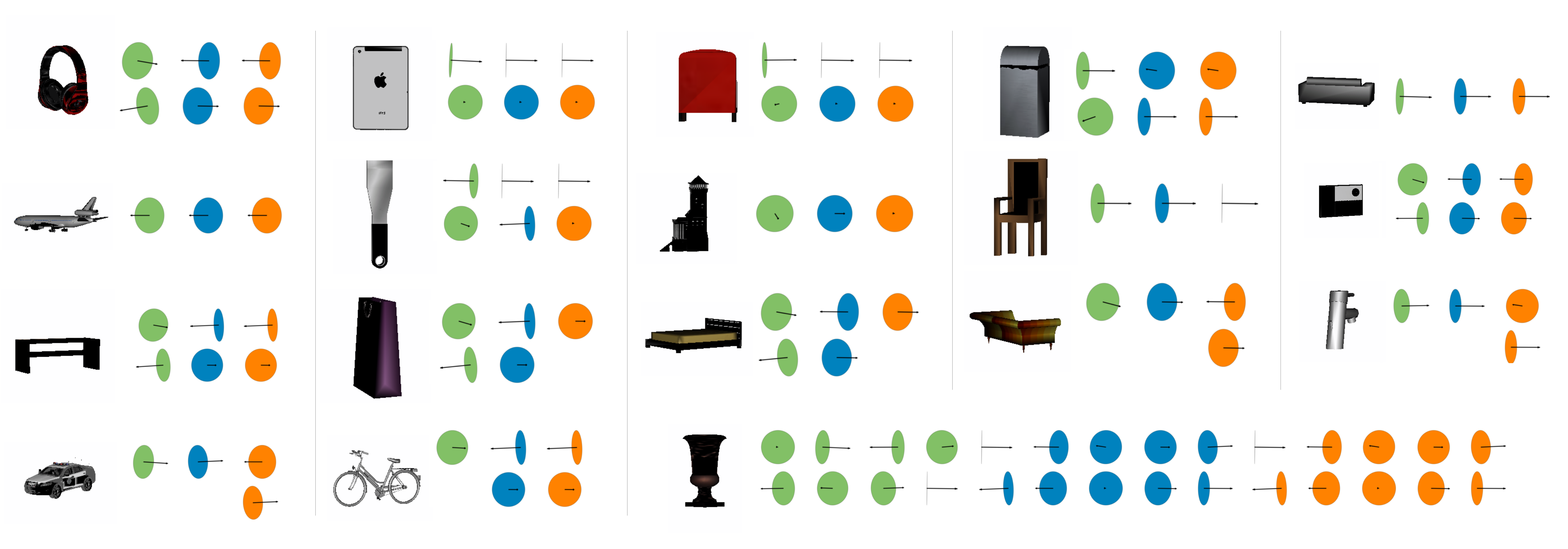}
\caption{Symmetry predictions for `Learned' and `Induced' settings for various test objects in our dataset. Each symmetry plane is visualized via a 3D circle parallel to the plane and an arrow denoting the normal to the plane. The green planes represent the ground-truth symmetries, the blue symmetry planes are predicted in the `Learned' setting and the orange symmetry planes are predicted under the `Induced' setting.}
\figlabel{symSynthRes}
\end{figure*}

\paragraph{Learning.}
Our data generation process is similar to the task of symmetry prediction described previously. We use a rendering engine $E$ which, given a shape model $S$ and a model pose $R$ yields a rendered image $E(S,R)$ and additionally provides a surface orientation image $\hat{N}(S,R)$. We first sample a category $c$ from training classes $\mathcal{C}$ and then a shape $S$. A random view $R$ is sampled from a view distribution $\mathcal{V}$ and the engine $E$ yields a rendering and normal image pair $(I, \hat{N})$. We then discretize $\hat{N}$ using the orientation bins above to obtain $N$ where $N(x,y) \in \{1,\cdots K\}$ is the orientation bin for the underlying surface. The pair $(I, N)$ forms a training sample for our learning system.

Given the training set constructed above, we train a CNN  to predict pixel-wise surface normals. Our architecture choice is motivated by recent methods that leverage CNNs to predict a dense pixel-wise output \eg semantic segmentation \cite{long_shelhamer_fcn} and image synthesis \cite{dosovitskiy2015inverting}. A common technique used in these architectures is to eschew fully connected layers common for image-level classification tasks and instead use multiple convolution layers followed by \textit{deconvolution} layers (reverse convolution with unpooling) to produce a dense pixelwise output. Let $C(k,s,o), D(k,s,o)$ denote a convolution layer with kernel size $k$, (downsampling(\textit{conv}) / upsampling(\textit{deconv})) stride $s$ and $o$ output channels and $P(k,s)$ represent a max-pooling layer with kernel size $k$ and stride $s$. Using the shorthand $C'(o)$ for $C(3,1,o)-C(3,1,o)-C(3,1,o)-P(2,2)$, our network architecture is $I-C'(64)-C'(128)-C'(256)-C'(512)-C'(512)-D(3,2,256)-D(3,2,128)-D(3,2,64)-D(3,2,K)$. The network above takes an input image and produces an output pixelwise log-probability distribution over the $K$ orientation bins. We minimize a softmax loss over the pixelwise log-probabilities predicted and train the CNN described using the Caffe framework. The convolutional layers are initialized using the VGG16 pretrained model for image classification \cite{simonyan2014very} and the deconvolution layers are initialized randomly. The architecture described produces a $113 \times 113$ spatial output given an input image of size $224 \times 224$ and this resolution allows our model to capture sharp discontinuities. In a similar spirit to symmetry orientation prediction, the category-agnostic formulation and learning of the surface orientation prediction  allow us to learn common representations to predict surface normals for novel objects.

\section{Experiments}
\seclabel{experiments}
Experiments were performed to investigate the following: 1) the performance of our symmetry and normal prediction systems and 2) their ability to generalize to novel unseen object categories. We first describe our experimental setup and then present results on symmetry detection and surface normal estimation. Finally, we show qualitative results on real world images in \figref{realRes}.

\begin{figure}[t!]
\includegraphics[width=0.45\textwidth]{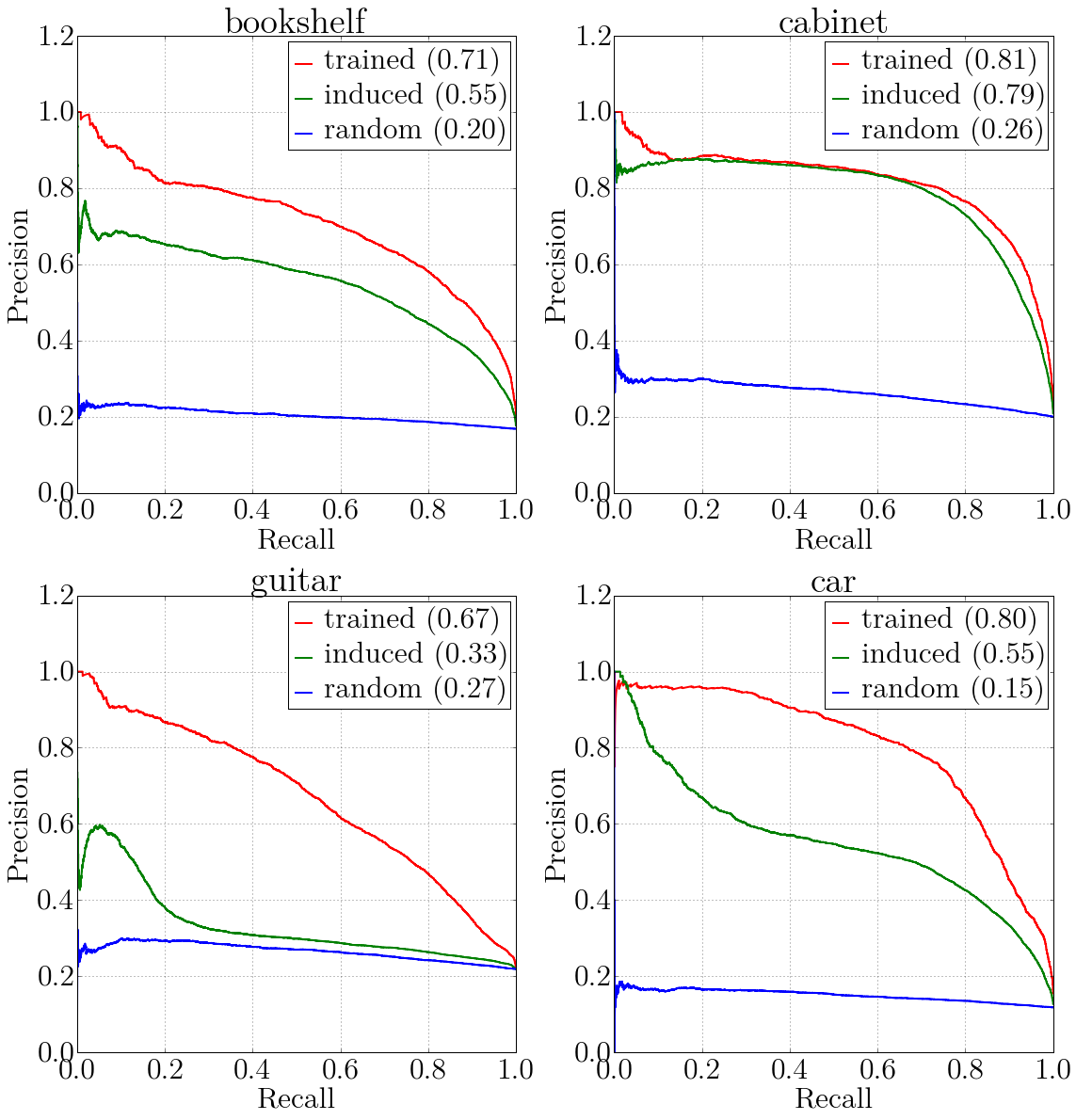}

\caption{Precision-Recall plots for symmetry detection under `Induced' and  `Learned' settings for representative classes.}
\figlabel{symPlots}
\end{figure}

\paragraph{Dataset.}
We use the ShapeNet \cite{shapenetWeb} dataset to download 3D models for objects corresponding to 57 object classes. These 3D models (collected from large scale 3D model repositories such as 3D Warehouse and Yobi3D) belong to object categories ranging from cars and buses to faucets and washers and form a varied set of commonly occurring rigid objects. We keep up to 200 models per object category (with a 75\%/25\% train/test split) and use 200 renderings for each 3D model in our training set to train the prediction systems previously described (totalling around 1.5 million images). In addition, we also sample equally from all classes for each training iteration in order to counter the class imbalance in the number of available models. Our testing set includes 3200 rendered images from each of the 57 object categories. For ease of reproducibility, we plan to make our train/test splits and code available.

\paragraph{Viewpoint Variability.}
The viewing angle for an object can be described using three euler angles - azimuth ($\phi \in (-180,180]$), elevation($\varphi  \in (-180,180]$) and cyclo-rotation($\psi  \in (-90,90]$). Objects, however, tend to follow certain view distributions (\eg we rarely see cars from the bottom).  In particular, the primary variation in viewing angle for objects in natural scenes is along the azimuth. To account for this, we sample views uniformly from a set of more \textit{natural} views $V_N = \{ \phi \in (-180,180] \}  \times \{ \varphi  \in [0,10] \} \times \{\psi  \in [0,0] \}$. It is, however, also important to handle objects seen from arbitrary views. We therefore also train and test our models under a more diverse view sampling from $V_D = \{ \phi \in (-180,180] \}  \times \{ \varphi  \in [0,50] \} \times \{\psi  \in [-30,30] \}$ to analyze the prediction and induction performance under more challenging settings.

\paragraph{Induction Splits.}
A primary aim of our experimental evaluation is to analyze the induction ability of our system across novel object classes. For this analysis, we randomly partitioned the object classes $\mathcal{C}$ in the ShapeNet dataset in two disjoint sets $\mathcal{C_A}$ and $\mathcal{C_B}$ - the categories in each set are listed in the appendix. For both the shape prediction tasks we study - normal and symmetry prediction, we train 3 models with the same hyper-parameters. One model is trained on the entire set of classes $\mathcal{C}$ and two models over $\mathcal{C_A}$ and $\mathcal{C_B}$ respectively. This allows us to empirically estimate the induction performance for a class $c \in (\mathcal{C_A} ~ \text{or} ~ \mathcal{C_B}) $ by comparing the performance of the systems trained over $\mathcal{C}$ and $(\mathcal{C_B} ~ \text{or} ~ \mathcal{C_A})$ respectively. In all the experiments described below, we report numbers under both the `Learned' and `Induced' settings. The `Learned' setting denotes the performance of our system when trained using \textit{all} object classes and the `Induced' setting indicates our performance when, for each object class we use the system trained on the set of classes $(\mathcal{C_A} ~ \text{or} ~ \mathcal{C_B}) $ \textit{not} containing the class under consideration. 

\begin{figure}
\includegraphics[width=0.48\textwidth]{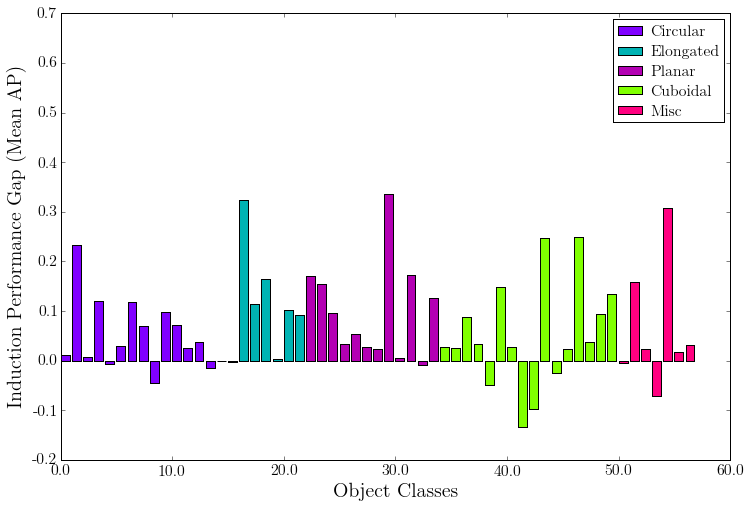}
\caption{Analysis of performance gap (in $AP_s^\theta$) for object categories between `Induced' and `Learned' settings for symmetry prediction.}
\figlabel{symAnalysis}
\end{figure}

\begin{table}
\centering
\setlength{\tabcolsep}{3pt}
\begin{tabular}{lcc|c}
\toprule
\textbf{Mean $AP_s^\theta$}  & \multicolumn{2}{c}{\textbf{Setting}} &\tabularnewline
\midrule
Viewpoint Sampling & Learned & Induced & Random \tabularnewline
\hline 
$V_N$ &  0.69  & 0.58 & 0.32 \tabularnewline
$V_D$ & 0.59  & 0.47 & 0.07 \tabularnewline
\bottomrule
\end{tabular}

\vspace{2mm}
\caption{Mean performance across classes for symmetry prediction.}
\tablelabel{symEval}
\end{table}

\begin{figure*}[t!]
  \centering
\includegraphics[width=0.9\textwidth]{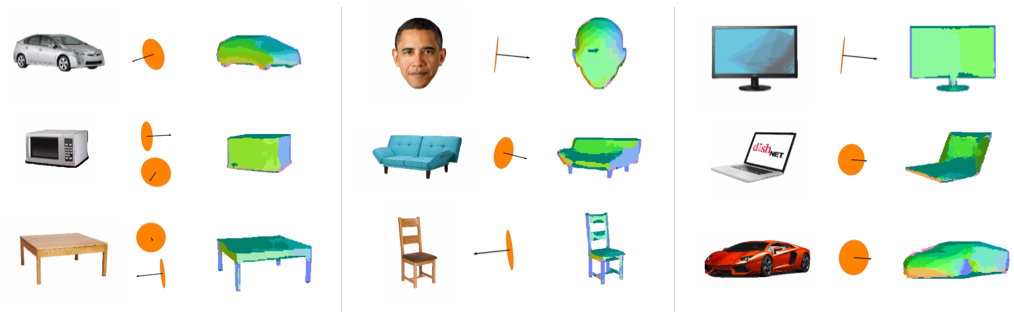}
\caption{Predicted symmetry orientations and surface normals for objects segmented out from real world images. The symmetries are shown using the convention in \figref{symSynthRes} and surface normals are mapped into RGB space via the mapping $X\rightarrow$\textcolor{blue}{B}, $Y\rightarrow$\textcolor{red}{R}, $Z\rightarrow$\textcolor{green}{G}.}
\figlabel{realRes}
\end{figure*}

\subsection{Symmetry Prediction}
\seclabel{symEval}

\paragraph{Evaluation Criterion.}
Since the task we address is not a standard one, we need to decide on evaluation metrics. In the related task of pose prediction, the common practice is to measure the deviation between predicted and annotated pose \cite{vpsKpsTulsianiM15, renderCNN}. Symmetries, however, do not lend themselves to a similar analysis because there can be multiple of them and consequently, a symmetry prediction system would yield multiple symmetry hypotheses with varying confidences. In that respect, our task perhaps has more in common with object detection - given an image with a variable number of symmetry planes (\cf objects), a prediction system outputs a few distinct hypotheses from the continuous space of plane orientations (\cf bounding box locations). We therefore adapt the standard object detection Average Precision (AP) metric for our task.

We propose $AP_s^\theta$ as a metric to  evaluate the performance of a symmetry prediction system. Given the ground-truth symmetries of an object $\mathcal{\hat{N}} = \{\hat{n}_i\}$ and predicted symmetry orientations $\mathcal{N} = \{n_j\}$ alongwith their probability scores $p_j$, a prediction $n \in \mathcal{N}$ is considered correct if $\exists i$ s.t $\Delta_s(n,\hat{n}_i)  \leq \theta$. Akin to the object detection setting, we also prevent double counting of ground-truth symmetries when matching a predicted symmetry. We vary the probability threshold for symmetry detection and consider all instances of a class together to obtain a point on the Precision-Recall curve. The $AP_s^\theta$ metric denotes the area under the above Precision-Recall curve.

\paragraph{Results.} 
We report the analysis of our system in \tableref{symEval}. We use $\theta = \frac{\pi}{18}$ for measuring the performance under the $AP_s^\theta$ metric. We report the performance of our systems for both view sampling settings $V_N$ and $V_D$. The system trained under the view sampling $V_N$ only classifies symmetry plane orientations among 10 possible horizontal directions whereas the system under $V_D$ setting predicts from among 60 possible orientations. We observe that the performance in the `Induced' setting, where we have not seen a single annotated object of the corresponding class, is comparable to the `Learned' setting with observed training examples. It is also encouraging that the results hold in the natural as well as diverse view sampling scenarios and that the `Induced' results are significantly better than an uninformed random baseline, thereby supporting our claim of the ability to generalize symmetry prediction across novel objects.

\paragraph{Analysis and Observations.} 
We manually grouped together object categories in coarse groups based on the shape of the typical bounding convex set. The resulting groups are indicated in \figref{symAnalysis} which also shows the performance gap, under the $V_N$ view sampling, between the `Learned' and `Induced' settings. We observe that the gap for a large fraction of the categories is low. In particular, we observe this trend for 'Circular' and 'Cuboidal' classes - this may  perhaps be a result of a large number of such classes being available for training and thus aiding generalization for novel objects of similar classes. 

We also show some predictions from our system in \figref{symSynthRes}. Both the induced as well as the trained system correctly predict most of the symmetries present in the objects. One of the primary error modes we observe in the 'Induced' setting is \textit{over-generalization} where the system confidently predicts symmetries in addition to the correct one for objects like motorbikes, rifles \etc, possibly on account of more commonly occurring classes with multiple symmetries. We show the performance under various settings in \figref{symPlots} for some representative classes. The first two are typical classes with strong generalization results whereas the performance on category `car' reduces significantly. 
%\newpage
\begin{figure*}
  \centering
  \includegraphics[width=\textwidth]{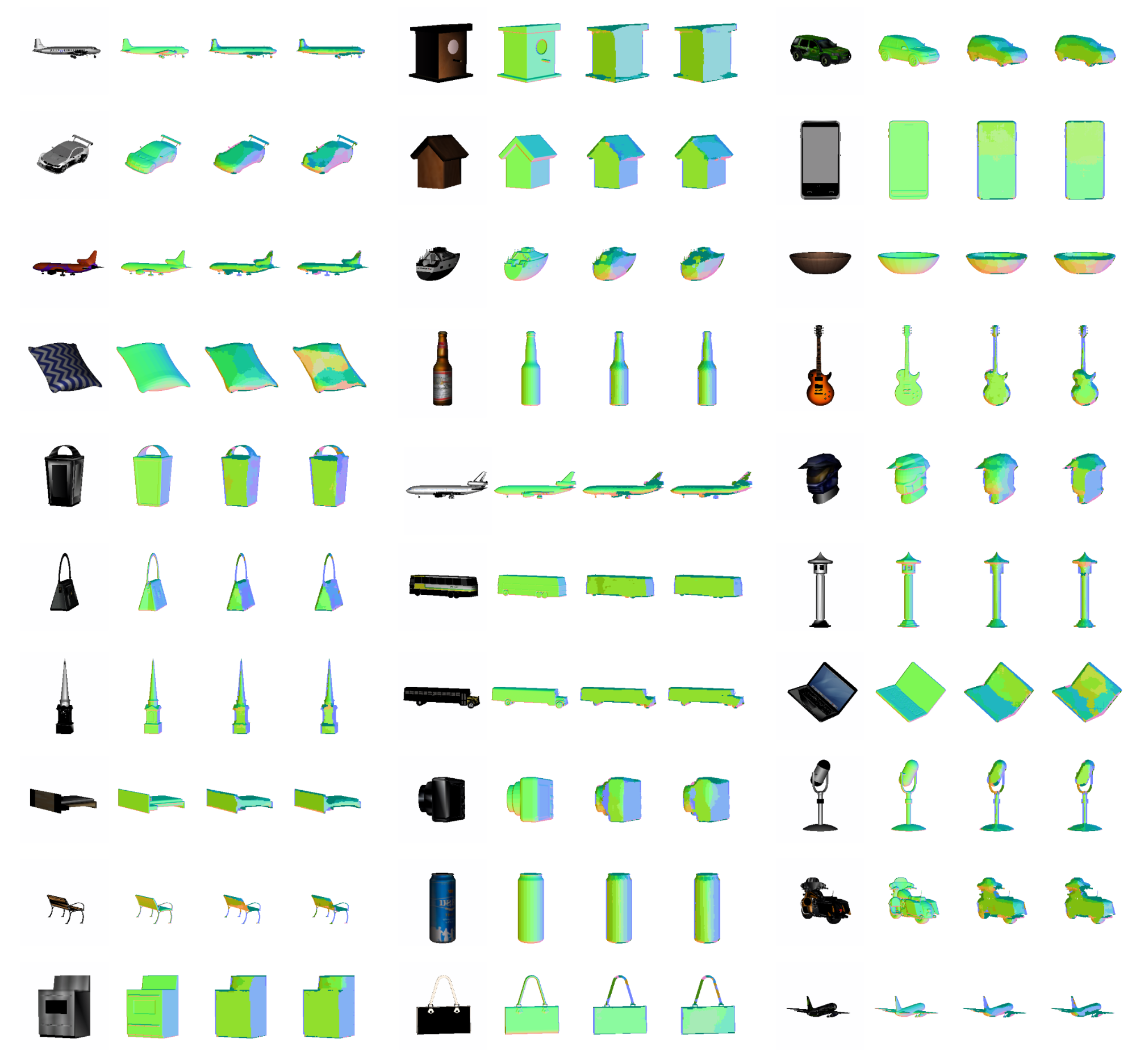} % COCO_train2014_000000352357.jpg

\caption{\figlabel{result_normals} Surface normal predictions using our learned and induced models on held out data. Each unit in the figure shows (from left to right) the image, its ground truth surface normals, surface normals predicted by our `Learned' model (trained on all classes) and surface normals predicted by our `Induced' model (trained on the subset of categories not containing this particular class). Surface normals are mapped into RGB space via the mapping $X\rightarrow$\textcolor{blue}{B}, $Y\rightarrow$\textcolor{red}{R}, $Z\rightarrow$\textcolor{green}{G}}
\end{figure*}

\subsection{Surface Normal Estimation}
% describe error metrics. discuss results and 
\paragraph{Evaluation Metrics.}
We follow the evaluation protocol from Fouhey \etal \cite{fouheyICCV13} and evaluate our predicted surface normals against the ground truth using 5 metrics - mean angular error, median angular error and the fraction of `good' pixels - pixels whose predicted normals lie within 11.25$^{\circ}$, 22.5$^{\circ}$ and $30^{\circ}$ of the ground truth normals respectively. All the above metrics are computed per object category and then reported below by averaging across the 57 classes. The mean and median angular error are computed across all object pixels per category (background is ignored) and so are the fraction of `good' pixels. We also report curves of fraction of `good' pixels vs. the angular threshold at which they are calculated and compute the area under the curve when the max angular threshold is 30$^{\circ}$.

\begin{table}[htb!]
\centering
\setlength{\tabcolsep}{3pt}
\begin{tabular}{lcc|cc}
\toprule
\textbf{Metrics}  & \multicolumn{2}{c}{\textbf{$V_N$}} & \multicolumn{2}{c}{\textbf{$V_D$}}\tabularnewline
\midrule
% & \textbf{L} & \textbf{I} & \textbf{L} & \textbf{I} \tabularnewline
& \textbf{Learned} & \textbf{Induced} & \textbf{Learned} & \textbf{Induced} \tabularnewline
Mean Error & 21.3 & 23.8 & 23.5 & 26.5\tabularnewline
Median Error & 12.7 & 14.7 & 14.8 & 17.1\tabularnewline
\midrule
\%GP $11.25^{\circ}$ & 50.4 & 45.7 & 41.6 & 37.1\tabularnewline
\%GP $22.5^{\circ}$ & 71.9 & 67.0 & 67.6 & 62.1\tabularnewline
\%GP $30.0^{\circ}$ & 77.8 & 73.5 & 74.7 & 69.5\tabularnewline
\bottomrule
\end{tabular}

\vspace{2mm}
\caption{Mean performance across classes for surface normal estimation under various view settings. Lower is better for the top half of the table and higher is better for the percent of `good' pixels metrics. Please refer to the text for more details on the metrics. }
\tablelabel{normal_results_table}
\end{table}

\paragraph{Results.}
The results for our surface normal prediction system(s) are shown in \tableref{normal_results_table}. As in the symmetry prediction task, we report results in both the $V_N$ and $V_D$ settings. For \textit{both} settings, the surface normal direction is discretized into 60 uniformly sampled bins on the hemisphere and the predicted labels are converted back into surface normals by looking up the orientation corresponding to the predicted bin. The `Learned' and `Induced' settings again refer to the experimental setups where the system was trained on all object classes and on the split of the dataset $(\mathcal{C_A} ~ \text{or} ~ \mathcal{C_B}) $ \textit{not} containing this object class respectively. It can be seen that our model achieves a pixel-wise median angular error rate of around $15^{\circ}$ and moreover the performance for the `Learned' and `Induced' settings are comparable, validating the claim that our surface normal prediction system generalizes to unseen object categories. This trend is visible across all error metrics as well as across viewpoint variation settings $V_N$ and $V_D$. 
\begin{figure}[htb!]
\includegraphics[width=0.48\textwidth]{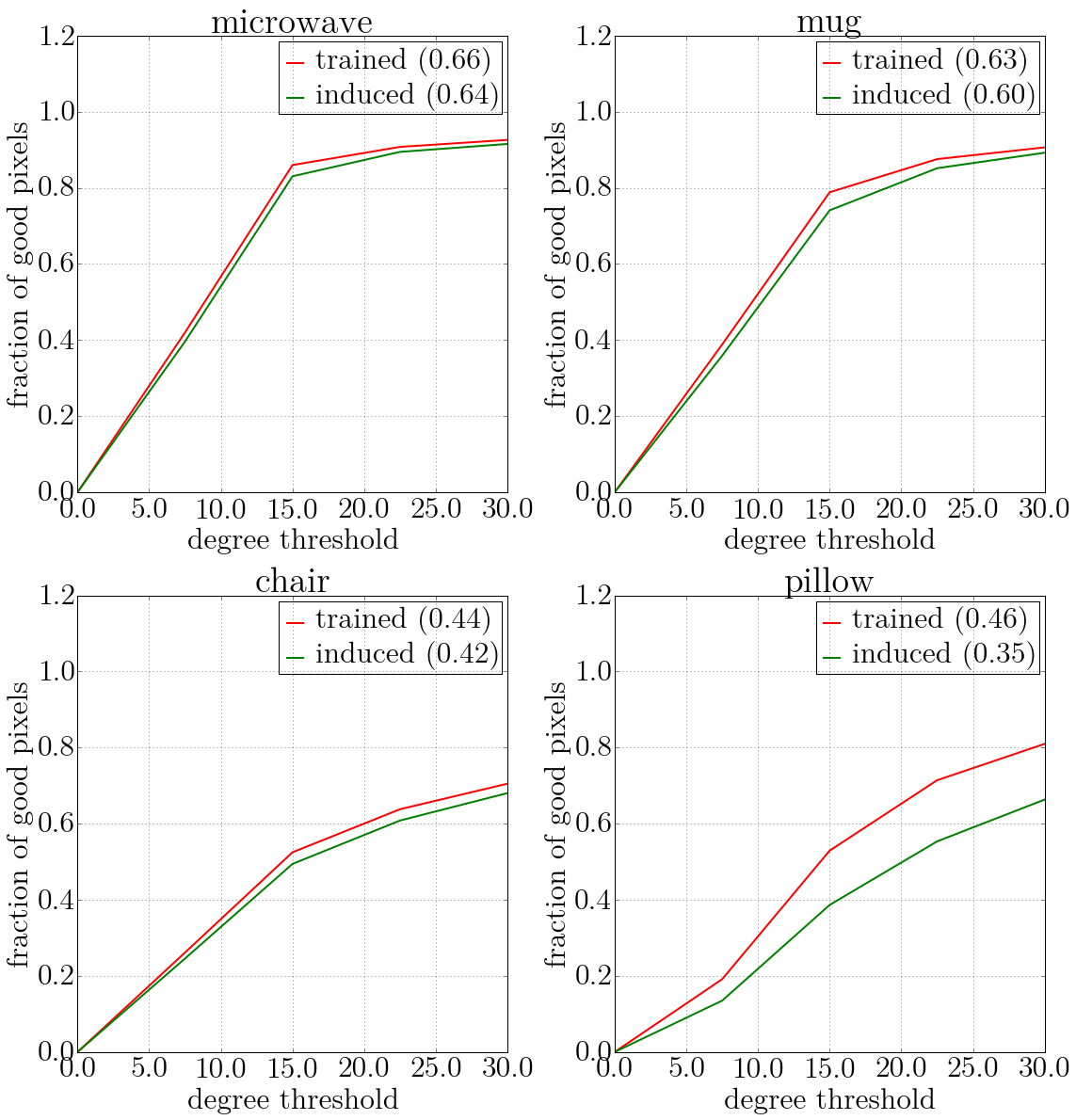}
\caption{Fraction of good pixels vs. degree threshold plots for surface normal prediction under `Induced' and  `Learned' settings for representative classes. The area under the curves are mentioned in the plot legends.}
\figlabel{normalPlots}
\end{figure}

\begin{figure}[htb!]
\includegraphics[width=0.45\textwidth]{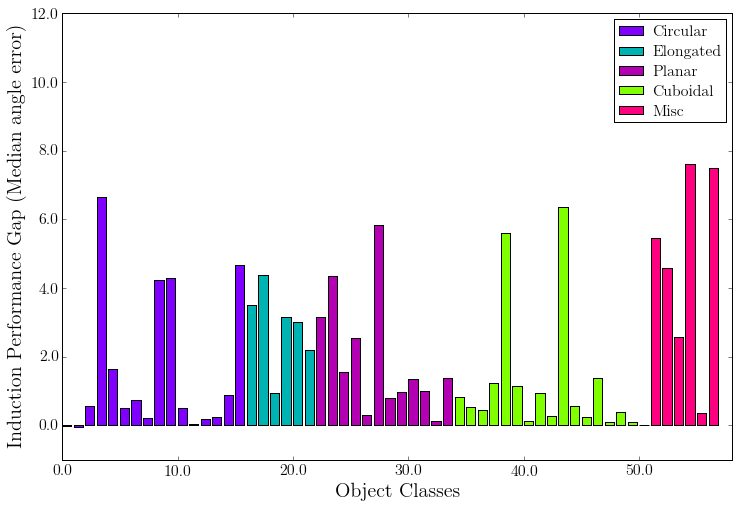}
\caption{Analysis of performance gap in (median angular error) between `Induced' and `Learned' settings for surface normal prediction.}
\figlabel{normalAnalysis}
\end{figure}

Some results from our surface normal predictor are shown in \figref{result_normals}. It can be seen that our system is able to reliably predict surface normals at a coarse level while also respecting discontinuities/edges. The major error modes for our system are fine structures which it is unable to handle owing to the large receptive fields in the middle convolutional layers in our architecture. \figref{normalAnalysis} shows the relative performance (in median angular error) of our `Learned' and `Induced' systems for various shape `super-categories' such as circular, cuboidal etc. It can be seen that the generalization works best for circular and cuboidal categories (consistent with the symmetry prediction experiments). We also show some plots for fraction of good pixels vs. angular threshold in \figref{normalPlots}. It can be seen that surface normals for `microwave' (cuboidal) and `mug' (cylindrical) generalize well whereas a non-standard shape such as `pillow' doesn't. `Chairs' on the other hand are overall worse off than other simpler categories but the coarse structures in them generalize well giving rise to similar curves for `Learned' and `Induced'. Detailed results and plots for symmetry prediction and surface normal estimation for all 57 object classes can be found in the appendix.

\section{Discussion}

Our results suggest that it is feasible to induce surface normals and 3D symmetry planes for objects from unfamiliar categories, by learning hierarchical feature extractors on a large-scale dataset of CAD model renderings. We also demonstrated that our learned models can operate on real images. The techniques we present here can in principle also be applied to predicting other shape properties such as local curvature, rotational symmetries \etc

Our approach connects modern representation learning approaches with the spirit of the pioneers in computer vision, that emphasized spatial vision and the understanding of shape. Should reconstruction be an input to classification, as for example Marr postulated \cite{marr1982vision}? Should it be the other way around? Or are both best handled as parallel processes, as in the dual stream hypothesis of neuroscience \cite{goodale1992separate}? We hope our approach would be useful for applications where shape understanding is important, including robotic perception and human-computer interaction.

\section*{Acknowledgements}
This work was supported in part by NSF Award IIS-1212798 and ONR MURI-N00014-10-1-0933. Shubham Tulsiani was supported by the Berkeley fellowship. Jo\~{a}o Carreira was supported by the Portuguese Science Foundation, FCT, under grant SFRH/BPD/84194/2012.  Qixing Huang thanks the gift awards from Adobe and Intel. We gratefully acknowledge NVIDIA corporation for GPU donations towards this research.

\bibliographystyle{ieee}

\bibliography{cvpr16geometry}

\clearpage
\setcounter{figure}{0}
\setcounter{table}{0}
\section*{Appendix}
%\subsection*{Splits}

\paragraph{Induction Splits.}
For the analysis of  the induction ability of our system across novel object classes, we randomly partitioned the object classes $\mathcal{C}$ in the ShapeNet dataset  \cite{shapenetWeb} in two disjoint sets $\mathcal{C_A}$ and $\mathcal{C_B}$. The categories in each set are listed in \tableref{inductionSplits}.

\begin{table}[htb!]
\centering
\setlength{\tabcolsep}{3pt}
\begin{tabular}{cc}
\toprule
$\mathcal{C_{A}}$ & $\mathcal{\mathcal{C_{B}}}$ \tabularnewline
\midrule
airplane & ashcan\tabularnewline
bathtub & bag\tabularnewline
bed & basket\tabularnewline
bicycle & bench\tabularnewline
bookshelf & birdhouse\tabularnewline
bottle & boat\tabularnewline
bowl & cabinet\tabularnewline
bus & camera\tabularnewline
can & cap\tabularnewline
clock & car\tabularnewline
computer keyboard & cellular telephone\tabularnewline
dishwasher & chair\tabularnewline
file & display\tabularnewline
loudspeaker & earphone\tabularnewline
mailbox & faucet\tabularnewline
microphone & guitar\tabularnewline
microwave & helmet\tabularnewline
mug & jar\tabularnewline
piano & knife\tabularnewline
pillow & lamp\tabularnewline
pistol & laptop\tabularnewline
pot & motorcycle\tabularnewline
printer & remote control\tabularnewline
skateboard & rifle\tabularnewline
stove & rocket\tabularnewline
table & sofa\tabularnewline
telephone & tower\tabularnewline
train & vessel\tabularnewline
 & washer\tabularnewline
\bottomrule
\end{tabular}
\vspace{2mm}
\caption{Induction Splits}
\tablelabel{inductionSplits}
\end{table}

\paragraph{Shape Groups.}
We provided additional analysis of our method by manually grouping together object categories in coarse groups based on the shape of the typical bounding convex set. The resulting groups used were as follows -
\begin{itemize}

\item \textbf{Circular :} ashcan, basket, bottle, bowl, can, cap, clock, helmet, jar, lamp, microphone, mug, pot, rocket, tower, washer

\item \textbf{Elongated :} computer keyboard, knife, piano, rifle, skateboard, train

\item \textbf{Planar :} airplane, bag, bench, bicycle, bookshelf, cellular telephone, display, file, laptop, motorcycle, pistol, remote control

\item \textbf{Cuboidal :} bathtub, bed, bus, cabinet, camera, car, chair, dishwasher, loudspeaker, mailbox, microwave, pillow, printer, sofa, stove, table

\item \textbf{Misc :} birdhouse, boat, earphone, faucet, guitar, telephone, vessel

\end{itemize}

\paragraph{Symmetry Prediction.}
We show the Precision-Recall plots for symmetry prediction for all classes  under the view sampling $V_n$ and $V_d$ in \figref{symPlotsVn} and \figref{symPlotsVd} respectively. The performance under  $AP_s^\theta$ metric is also reported in \tableref{symResults}.

\begin{figure*}[t!]
\begin{center}

\includegraphics[height=0.9\textheight]{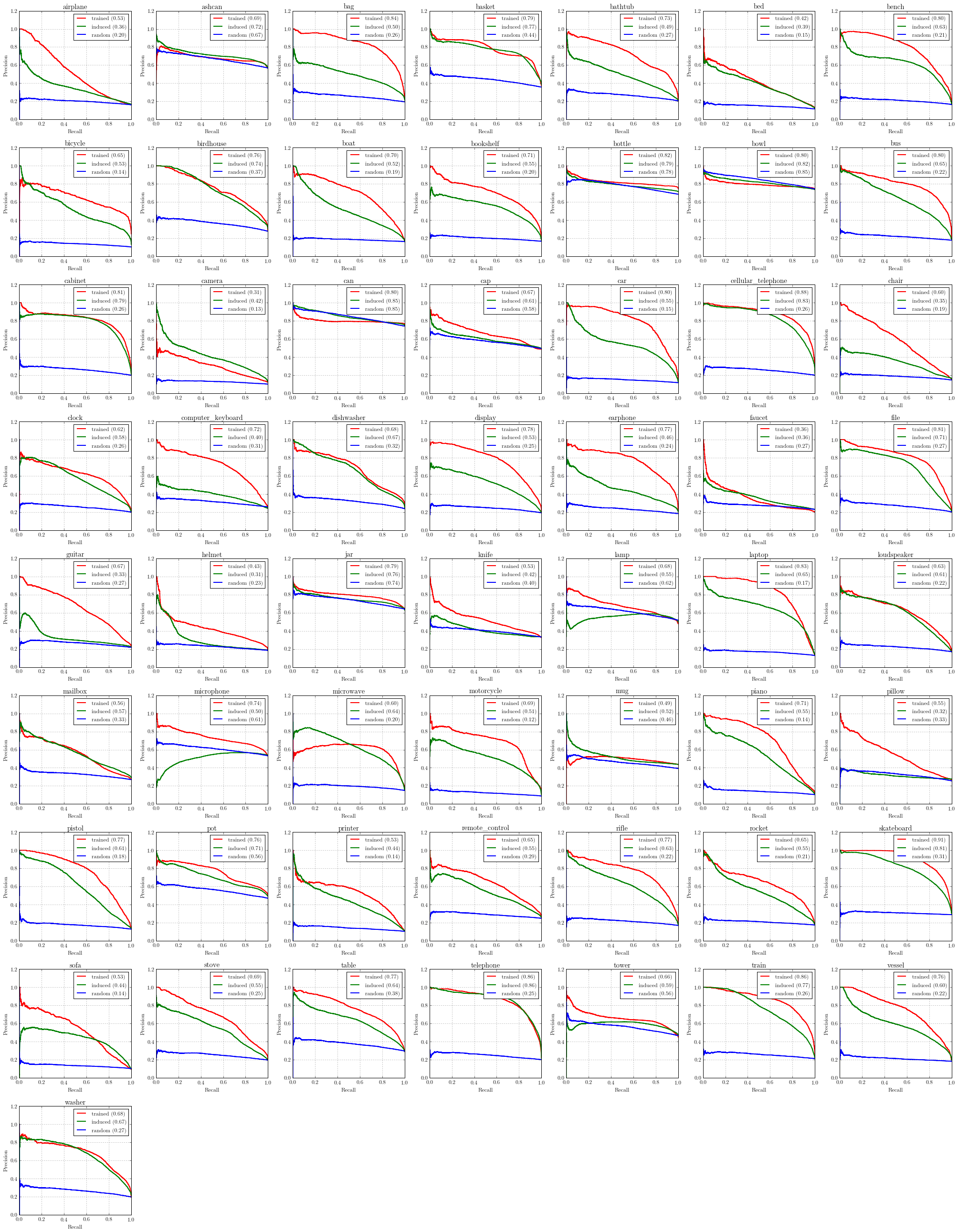}
\caption{Precision-Recall plots for symmetry detection under `Induced' and  `Learned' settings and $V_n$ view sampling for all classes.}
\figlabel{symPlotsVn}
\end{center}
\end{figure*}

\begin{figure*}[t!]
\begin{center}

\includegraphics[height=0.9\textheight]{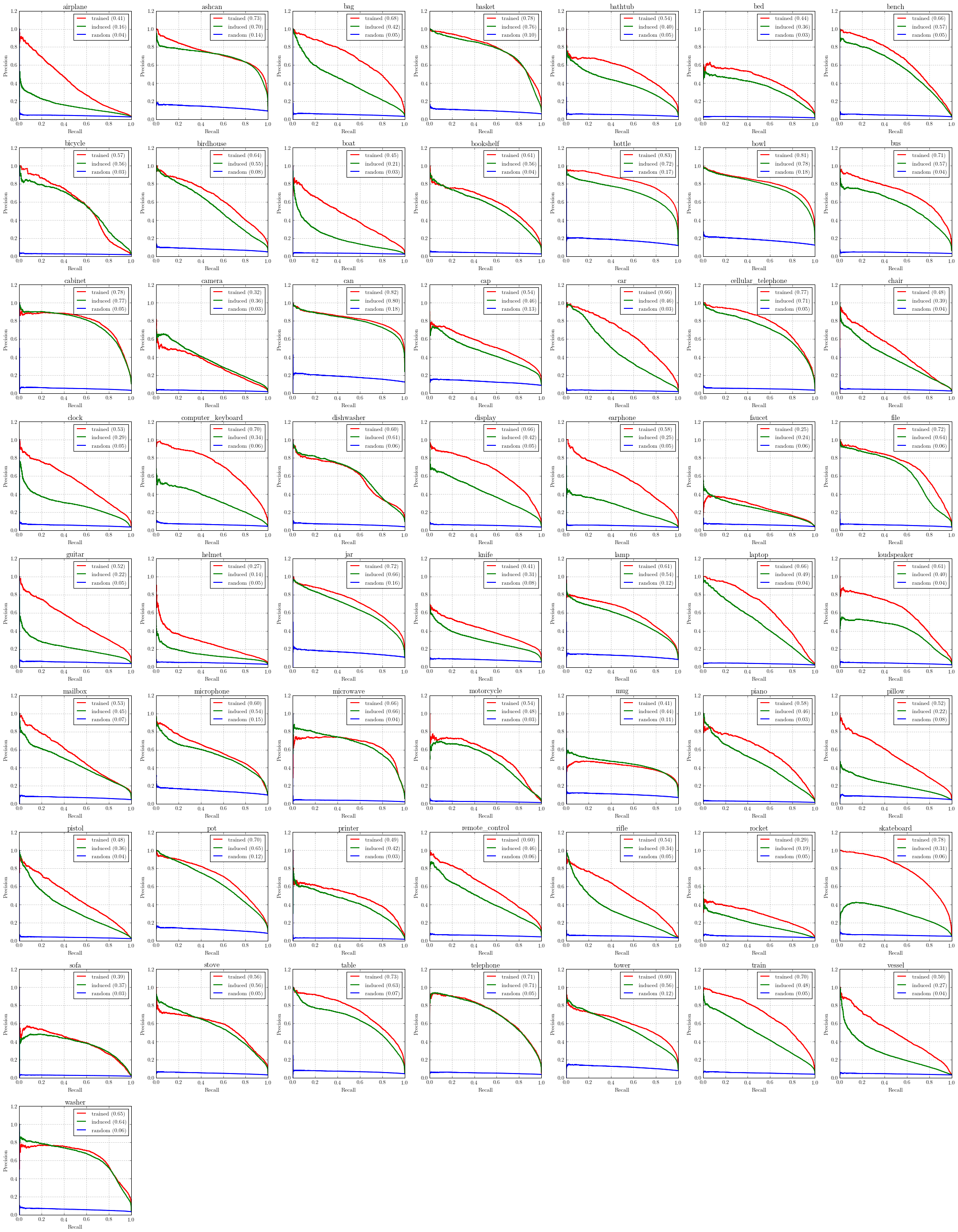}
\caption{Precision-Recall plots for symmetry detection under `Induced' and  `Learned' settings and $V_d$ view sampling for all classes.}
\figlabel{symPlotsVd}
\end{center}
\end{figure*}

\begin{table*}[htb!]
\centering
\footnotesize
\setlength{\tabcolsep}{3pt}
\begin{tabular}{lccc|ccc}
\toprule
\textbf{View Setting}  & \multicolumn{3}{c}{\textbf{$V_D$}} & \multicolumn{3}{c}{\textbf{$V_N$}}\tabularnewline
\midrule
  & Learned & Induced & Random & Learned & Induced & Random \tabularnewline
\midrule
airplane & 0.41 & 0.16 & 0.04 & 0.53 & 0.36 & 0.20\tabularnewline
ashcan & 0.73 & 0.70 & 0.14 & 0.69 & 0.72 & 0.67\tabularnewline
bag & 0.68 & 0.42 & 0.05 & 0.84 & 0.50 & 0.26\tabularnewline
basket & 0.78 & 0.76 & 0.10 & 0.79 & 0.77 & 0.44\tabularnewline
bathtub & 0.54 & 0.40 & 0.05 & 0.73 & 0.49 & 0.27\tabularnewline
bed & 0.44 & 0.36 & 0.03 & 0.42 & 0.39 & 0.15\tabularnewline
bench & 0.66 & 0.57 & 0.05 & 0.80 & 0.63 & 0.21\tabularnewline
bicycle & 0.57 & 0.56 & 0.03 & 0.65 & 0.53 & 0.14\tabularnewline
birdhouse & 0.64 & 0.55 & 0.08 & 0.76 & 0.74 & 0.37\tabularnewline
boat & 0.45 & 0.21 & 0.03 & 0.70 & 0.52 & 0.19\tabularnewline
bookshelf & 0.61 & 0.56 & 0.04 & 0.71 & 0.55 & 0.20\tabularnewline
bottle & 0.83 & 0.72 & 0.17 & 0.82 & 0.79 & 0.78\tabularnewline
bowl & 0.81 & 0.78 & 0.18 & 0.80 & 0.82 & 0.85\tabularnewline
bus & 0.71 & 0.57 & 0.04 & 0.80 & 0.65 & 0.22\tabularnewline
cabinet & 0.78 & 0.77 & 0.05 & 0.81 & 0.79 & 0.26\tabularnewline
camera & 0.32 & 0.36 & 0.03 & 0.31 & 0.42 & 0.13\tabularnewline
can & 0.82 & 0.80 & 0.18 & 0.80 & 0.85 & 0.85\tabularnewline
cap & 0.54 & 0.46 & 0.13 & 0.67 & 0.61 & 0.58\tabularnewline
car & 0.66 & 0.46 & 0.03 & 0.80 & 0.55 & 0.15\tabularnewline
cellular\_telephone & 0.77 & 0.71 & 0.05 & 0.88 & 0.83 & 0.26\tabularnewline
chair & 0.48 & 0.39 & 0.04 & 0.60 & 0.35 & 0.19\tabularnewline
clock & 0.53 & 0.29 & 0.05 & 0.62 & 0.58 & 0.26\tabularnewline
computer\_keyboard & 0.70 & 0.34 & 0.06 & 0.72 & 0.40 & 0.31\tabularnewline
dishwasher & 0.60 & 0.61 & 0.06 & 0.68 & 0.67 & 0.32\tabularnewline
display & 0.66 & 0.42 & 0.05 & 0.78 & 0.53 & 0.25\tabularnewline
earphone & 0.58 & 0.25 & 0.05 & 0.77 & 0.46 & 0.24\tabularnewline
faucet & 0.25 & 0.24 & 0.06 & 0.36 & 0.36 & 0.27\tabularnewline
file & 0.72 & 0.64 & 0.06 & 0.81 & 0.71 & 0.27\tabularnewline
guitar & 0.52 & 0.22 & 0.05 & 0.67 & 0.33 & 0.27\tabularnewline
helmet & 0.27 & 0.14 & 0.05 & 0.43 & 0.31 & 0.23\tabularnewline
jar & 0.72 & 0.66 & 0.16 & 0.79 & 0.76 & 0.74\tabularnewline
knife & 0.41 & 0.31 & 0.08 & 0.53 & 0.42 & 0.40\tabularnewline
lamp & 0.61 & 0.54 & 0.12 & 0.68 & 0.55 & 0.62\tabularnewline
laptop & 0.66 & 0.49 & 0.04 & 0.83 & 0.65 & 0.17\tabularnewline
loudspeaker & 0.61 & 0.40 & 0.04 & 0.63 & 0.61 & 0.22\tabularnewline
mailbox & 0.53 & 0.45 & 0.07 & 0.56 & 0.57 & 0.33\tabularnewline
microphone & 0.60 & 0.54 & 0.15 & 0.74 & 0.50 & 0.61\tabularnewline
microwave & 0.66 & 0.66 & 0.04 & 0.60 & 0.64 & 0.20\tabularnewline
motorcycle & 0.54 & 0.48 & 0.03 & 0.69 & 0.51 & 0.12\tabularnewline
mug & 0.41 & 0.44 & 0.11 & 0.49 & 0.52 & 0.46\tabularnewline
piano & 0.58 & 0.46 & 0.03 & 0.71 & 0.55 & 0.14\tabularnewline
pillow & 0.52 & 0.22 & 0.08 & 0.55 & 0.32 & 0.33\tabularnewline
pistol & 0.48 & 0.36 & 0.04 & 0.77 & 0.61 & 0.18\tabularnewline
pot & 0.70 & 0.65 & 0.12 & 0.76 & 0.71 & 0.56\tabularnewline
printer & 0.49 & 0.42 & 0.03 & 0.53 & 0.44 & 0.14\tabularnewline
remote\_control & 0.60 & 0.46 & 0.06 & 0.65 & 0.55 & 0.29\tabularnewline
rifle & 0.54 & 0.34 & 0.05 & 0.77 & 0.63 & 0.22\tabularnewline
rocket & 0.29 & 0.19 & 0.05 & 0.65 & 0.55 & 0.21\tabularnewline
skateboard & 0.78 & 0.31 & 0.06 & 0.91 & 0.81 & 0.31\tabularnewline
sofa & 0.39 & 0.37 & 0.03 & 0.53 & 0.44 & 0.14\tabularnewline
stove & 0.56 & 0.56 & 0.05 & 0.69 & 0.55 & 0.25\tabularnewline
table & 0.73 & 0.63 & 0.07 & 0.77 & 0.64 & 0.38\tabularnewline
telephone & 0.71 & 0.71 & 0.05 & 0.86 & 0.86 & 0.25\tabularnewline
tower & 0.60 & 0.56 & 0.12 & 0.66 & 0.59 & 0.56\tabularnewline
train & 0.70 & 0.48 & 0.05 & 0.86 & 0.77 & 0.26\tabularnewline
vessel & 0.50 & 0.27 & 0.04 & 0.76 & 0.60 & 0.22\tabularnewline
washer & 0.65 & 0.64 & 0.06 & 0.68 & 0.67 & 0.27\tabularnewline
\bottomrule
\normalsize
\end{tabular}

\vspace{2mm}
\caption{Performance across classes for symmetry prediction under $AP_s^\theta$ metric. }
\tablelabel{symResults}
\end{table*}
%\includepdf{cvpr16geometrySupp.pdf}
\paragraph{Normal Estimation.}
We show the performance plots for normal estimation for all classes  under the view sampling $V_n$ and $V_d$ in \figref{normPlotsVn} and \figref{normPlotsVd} respectively. The performance under  various metrics under the view sampling $V_n$ and $V_d$  are reported in \tableref{normalResVn} and \tableref{normalResVd} respectively.

\begin{figure*}[t!]
\begin{center}

\includegraphics[height=0.9\textheight]{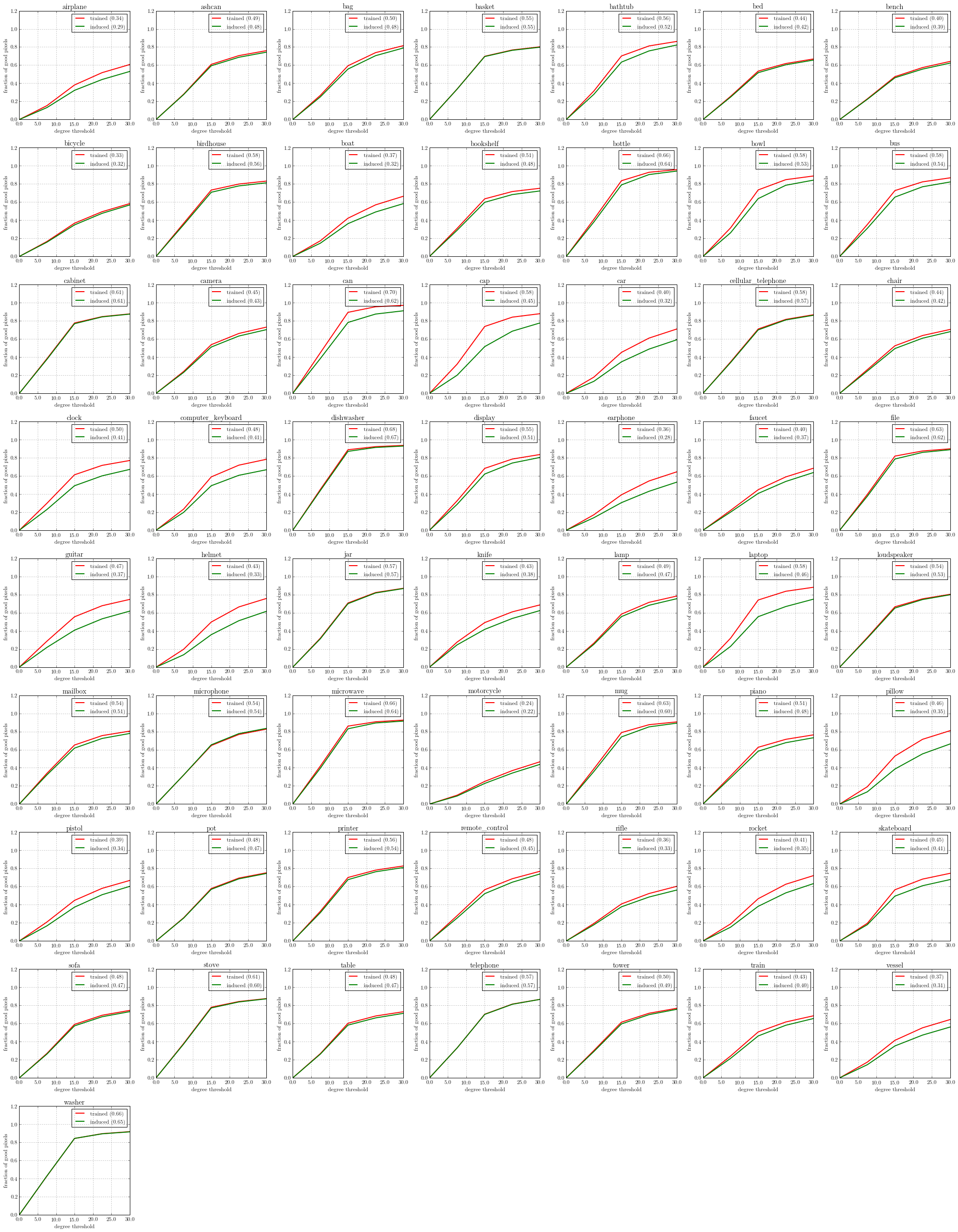}
\caption{Fraction of good pixels vs. degree threshold plots for surface normal prediction   under `Induced' and  `Learned' settings and $V_n$ view sampling for all classes. The area under the curves are mentioned in the plot legends.}
\figlabel{normPlotsVn}
\end{center}
\end{figure*}

\begin{figure*}[t!]
\begin{center}

\includegraphics[height=0.9\textheight]{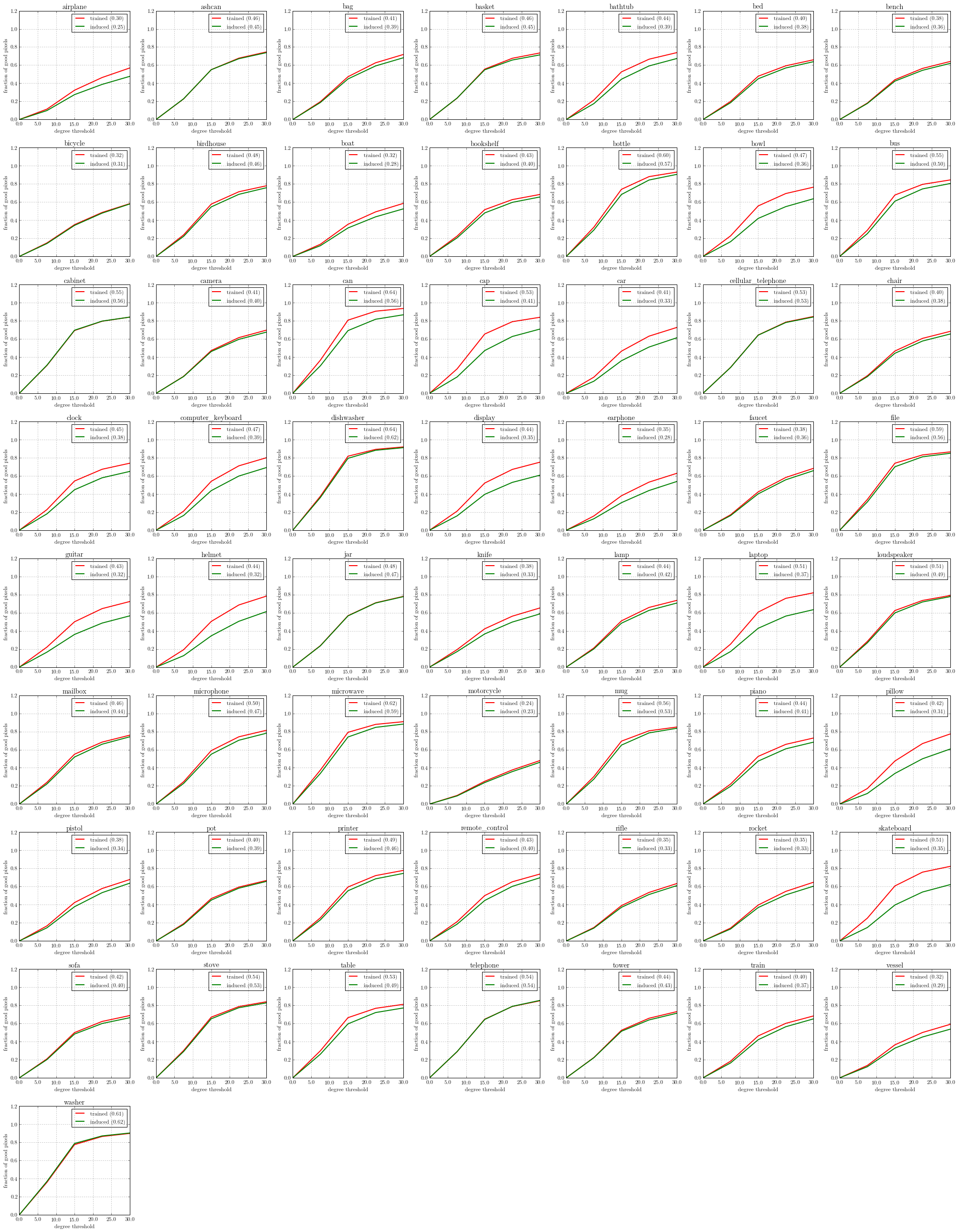}
\caption{Fraction of good pixels vs. degree threshold plots for surface normal prediction   under `Induced' and  `Learned' settings and $V_d$ view sampling for all classes. The area under the curves are mentioned in the plot legends.}
\figlabel{normPlotsVd}
\end{center}
\end{figure*}

\begin{table*}[htb!]
\centering
\footnotesize
\setlength{\tabcolsep}{3pt}
\begin{tabular}{lccccccc|cccc}
\toprule
\textbf{Metrics} &  & \textbf{\%GP $11.25^{\circ} $} & & \textbf{\%GP $22.5^{\circ}$} & & \textbf{\%GP $30.0^{\circ}$} & & & \textbf{Mean Error} & & \textbf{Median Error} \tabularnewline
\midrule
airplane &  & 29.3 (25.2) &  & 51.7 (44.2) &  & 60.7 (53.0) &  &  & 30.9 (34.7) &  & 21.4 (27.2)\tabularnewline
ashcan &  & 50.0 (48.8) &  & 70.5 (68.7) &  & 76.0 (74.3) &  &  & 21.9 (22.9) &  & 11.3 (11.5)\tabularnewline
bag &  & 46.2 (43.0) &  & 73.8 (70.4) &  & 81.5 (78.9) &  &  & 19.2 (20.5) &  & 12.2 (13.2)\tabularnewline
basket &  & 60.0 (59.7) &  & 76.8 (76.4) &  & 80.2 (79.8) &  &  & 20.1 (20.3) &  & 9.4 (9.5)\tabularnewline
bathtub &  & 57.6 (50.8) &  & 81.2 (75.7) &  & 86.1 (82.2) &  &  & 17.2 (19.6) &  & 9.8 (11.1)\tabularnewline
bed &  & 44.8 (43.2) &  & 61.8 (60.4) &  & 66.8 (65.6) &  &  & 28.6 (29.5) &  & 13.2 (14.0)\tabularnewline
bench &  & 40.0 (39.1) &  & 57.3 (55.6) &  & 64.1 (62.3) &  &  & 29.2 (30.3) &  & 16.7 (17.7)\tabularnewline
bicycle &  & 28.6 (27.1) &  & 49.4 (47.6) &  & 58.3 (56.8) &  &  & 31.9 (32.7) &  & 23.0 (24.4)\tabularnewline
birdhouse &  & 62.3 (59.6) &  & 79.9 (77.8) &  & 83.1 (81.2) &  &  & 18.4 (19.5) &  & 9.2 (9.6)\tabularnewline
boat &  & 32.0 (27.6) &  & 56.9 (49.0) &  & 66.3 (58.2) &  &  & 28.2 (32.4) &  & 18.6 (23.2)\tabularnewline
bookshelf &  & 54.0 (50.6) &  & 71.6 (68.2) &  & 75.2 (72.2) &  &  & 23.7 (25.6) &  & 10.3 (11.1)\tabularnewline
bottle &  & 69.5 (64.2) &  & 92.8 (90.3) &  & 95.7 (94.2) &  &  & 11.0 (12.2) &  & 8.5 (9.0)\tabularnewline
bowl &  & 58.4 (48.8) &  & 84.7 (78.5) &  & 88.7 (84.1) &  &  & 15.5 (18.8) &  & 9.9 (11.5)\tabularnewline
bus &  & 61.2 (53.2) &  & 82.3 (76.9) &  & 86.7 (82.1) &  &  & 16.3 (19.5) &  & 9.3 (10.5)\tabularnewline
cabinet &  & 67.5 (66.1) &  & 84.6 (84.3) &  & 87.6 (87.4) &  &  & 15.2 (15.5) &  & 8.7 (8.9)\tabularnewline
camera &  & 43.1 (40.9) &  & 66.1 (63.2) &  & 73.1 (70.3) &  &  & 24.0 (25.6) &  & 13.6 (14.5)\tabularnewline
can &  & 76.1 (65.1) &  & 95.5 (87.5) &  & 97.1 (91.0) &  &  & 9.8 (13.2) &  & 8.0 (8.9)\tabularnewline
cap &  & 58.6 (38.3) &  & 84.1 (68.6) &  & 87.9 (77.6) &  &  & 15.9 (22.2) &  & 9.8 (14.5)\tabularnewline
car &  & 33.5 (25.4) &  & 61.0 (48.8) &  & 70.9 (59.0) &  &  & 25.3 (31.8) &  & 16.9 (23.3)\tabularnewline
cellular\_telephone &  & 59.3 (58.6) &  & 81.6 (80.9) &  & 86.7 (86.1) &  &  & 16.3 (16.7) &  & 9.5 (9.7)\tabularnewline
chair &  & 43.2 (40.5) &  & 63.8 (60.8) &  & 70.5 (68.0) &  &  & 25.9 (27.3) &  & 13.9 (15.3)\tabularnewline
clock &  & 50.2 (39.3) &  & 71.6 (60.0) &  & 76.9 (67.1) &  &  & 21.6 (26.9) &  & 11.2 (15.4)\tabularnewline
computer\_keyboard &  & 47.3 (40.0) &  & 71.8 (60.6) &  & 78.3 (66.8) &  &  & 21.9 (29.2) &  & 11.9 (15.4)\tabularnewline
dishwasher &  & 79.6 (76.9) &  & 92.2 (91.3) &  & 93.6 (92.9) &  &  & 11.6 (12.2) &  & 7.9 (8.0)\tabularnewline
display &  & 56.0 (49.6) &  & 78.5 (74.2) &  & 83.5 (80.3) &  &  & 18.0 (20.4) &  & 10.0 (11.4)\tabularnewline
earphone &  & 29.2 (23.1) &  & 54.4 (43.0) &  & 64.4 (53.0) &  &  & 28.9 (34.6) &  & 20.0 (27.6)\tabularnewline
faucet &  & 35.6 (32.4) &  & 58.8 (53.8) &  & 68.4 (63.6) &  &  & 26.3 (28.6) &  & 17.5 (20.1)\tabularnewline
file &  & 71.1 (66.9) &  & 87.4 (85.9) &  & 89.8 (88.8) &  &  & 14.2 (15.1) &  & 8.6 (8.9)\tabularnewline
guitar &  & 45.0 (32.8) &  & 67.8 (53.3) &  & 74.7 (61.8) &  &  & 22.9 (29.7) &  & 12.8 (20.3)\tabularnewline
helmet &  & 36.6 (25.7) &  & 66.4 (51.2) &  & 75.8 (61.6) &  &  & 22.6 (30.2) &  & 15.1 (21.7)\tabularnewline
jar &  & 56.5 (55.5) &  & 82.3 (81.8) &  & 87.0 (86.7) &  &  & 16.2 (16.5) &  & 10.1 (10.3)\tabularnewline
knife &  & 40.4 (34.6) &  & 61.0 (53.7) &  & 68.6 (62.4) &  &  & 25.5 (29.1) &  & 15.5 (19.9)\tabularnewline
lamp &  & 46.9 (44.7) &  & 71.5 (68.3) &  & 78.4 (75.5) &  &  & 21.1 (22.6) &  & 12.1 (12.9)\tabularnewline
laptop &  & 61.3 (44.7) &  & 83.7 (66.9) &  & 88.1 (75.0) &  &  & 15.7 (23.4) &  & 9.6 (12.7)\tabularnewline
loudspeaker &  & 56.8 (55.2) &  & 75.4 (74.5) &  & 80.4 (79.9) &  &  & 19.4 (19.8) &  & 9.9 (10.1)\tabularnewline
mailbox &  & 54.8 (51.7) &  & 75.5 (72.4) &  & 80.4 (77.9) &  &  & 19.6 (21.1) &  & 10.2 (10.8)\tabularnewline
microphone &  & 52.1 (52.5) &  & 76.7 (77.7) &  & 82.6 (83.5) &  &  & 18.7 (18.4) &  & 10.7 (10.7)\tabularnewline
microwave &  & 76.6 (71.7) &  & 90.7 (89.4) &  & 92.6 (91.5) &  &  & 12.3 (13.3) &  & 8.1 (8.4)\tabularnewline
motorcycle &  & 18.1 (16.4) &  & 36.9 (34.2) &  & 46.6 (43.8) &  &  & 38.1 (39.7) &  & 32.9 (35.5)\tabularnewline
mug &  & 65.7 (60.5) &  & 87.5 (85.1) &  & 90.6 (89.2) &  &  & 13.9 (15.0) &  & 8.8 (9.4)\tabularnewline
piano &  & 52.8 (48.8) &  & 71.4 (67.7) &  & 76.3 (73.1) &  &  & 22.4 (24.5) &  & 10.6 (11.5)\tabularnewline
pillow &  & 37.7 (27.3) &  & 71.4 (55.3) &  & 81.0 (66.3) &  &  & 20.3 (27.7) &  & 14.2 (19.8)\tabularnewline
pistol &  & 35.4 (28.8) &  & 58.0 (51.0) &  & 66.7 (60.2) &  &  & 27.1 (31.0) &  & 17.5 (21.9)\tabularnewline
pot &  & 45.9 (45.2) &  & 69.3 (68.5) &  & 75.2 (74.4) &  &  & 22.6 (23.0) &  & 12.3 (12.5)\tabularnewline
printer &  & 59.1 (56.1) &  & 78.1 (76.3) &  & 82.7 (80.9) &  &  & 18.6 (19.8) &  & 9.6 (10.1)\tabularnewline
remote\_control &  & 45.9 (42.0) &  & 68.7 (64.8) &  & 76.9 (73.9) &  &  & 21.7 (23.3) &  & 12.5 (14.1)\tabularnewline
rifle &  & 32.9 (30.4) &  & 52.3 (48.5) &  & 60.2 (56.1) &  &  & 31.0 (33.7) &  & 20.7 (23.9)\tabularnewline
rocket &  & 35.4 (28.9) &  & 62.6 (53.0) &  & 72.0 (63.2) &  &  & 24.5 (28.9) &  & 16.4 (20.7)\tabularnewline
skateboard &  & 46.6 (40.6) &  & 68.2 (60.9) &  & 74.6 (67.7) &  &  & 23.4 (28.1) &  & 12.3 (15.3)\tabularnewline
sofa &  & 48.1 (46.8) &  & 69.1 (67.4) &  & 74.4 (73.0) &  &  & 24.0 (24.8) &  & 11.7 (12.1)\tabularnewline
stove &  & 67.6 (66.2) &  & 84.2 (83.7) &  & 87.5 (87.2) &  &  & 15.3 (15.6) &  & 8.8 (8.9)\tabularnewline
table &  & 52.6 (50.8) &  & 68.2 (66.1) &  & 72.9 (71.1) &  &  & 24.6 (25.8) &  & 10.4 (10.9)\tabularnewline
telephone &  & 58.1 (57.6) &  & 81.4 (81.2) &  & 86.6 (86.6) &  &  & 16.5 (16.5) &  & 9.8 (9.8)\tabularnewline
tower &  & 51.1 (49.2) &  & 71.3 (69.7) &  & 76.7 (75.6) &  &  & 21.8 (22.4) &  & 11.0 (11.5)\tabularnewline
train &  & 41.1 (36.9) &  & 61.6 (58.0) &  & 68.4 (65.4) &  &  & 26.5 (28.6) &  & 14.7 (16.9)\tabularnewline
vessel &  & 31.8 (27.1) &  & 55.2 (47.2) &  & 64.3 (56.1) &  &  & 29.1 (33.7) &  & 19.3 (24.8)\tabularnewline
washer &  & 74.5 (74.9) &  & 89.4 (89.3) &  & 91.7 (91.4) &  &  & 12.8 (12.9) &  & 8.2 (8.2)\tabularnewline
\bottomrule
\normalsize
\end{tabular}

\vspace{2mm}
\caption{Surface normal estimation performance for all classes under the $V_n$ view sampling. The numbers outside the parenthesis denote the `Learned' setting and the evaluation in the `Induced' setting is reported in the parenthesis. Higher is better for the first three percent of `good' pixels metrics and lower is better for last two three error metrics.}
\tablelabel{normalResVn}
\end{table*}

\begin{table*}[htb!]
\centering
\footnotesize
\setlength{\tabcolsep}{3pt}
\begin{tabular}{lccccccc|cccc}
\toprule
\textbf{Metrics} &  & \textbf{\%GP $11.25^{\circ} $} & & \textbf{\%GP $22.5^{\circ}$} & & \textbf{\%GP $30.0^{\circ}$} & & & \textbf{Mean Error} & & \textbf{Median Error} \tabularnewline
\midrule
airplane &  & 23.2 (20.0) &  & 46.4 (38.8) &  & 56.9 (47.6) &  &  & 32.7 (37.7) &  & 24.9 (32.3)\tabularnewline
ashcan &  & 42.0 (42.3) &  & 67.7 (67.1) &  & 74.6 (73.8) &  &  & 23.2 (23.6) &  & 13.3 (13.3)\tabularnewline
bag &  & 35.6 (34.0) &  & 62.6 (59.1) &  & 71.7 (68.1) &  &  & 24.9 (26.9) &  & 16.1 (17.2)\tabularnewline
basket &  & 42.9 (42.9) &  & 67.6 (65.7) &  & 73.5 (71.3) &  &  & 24.8 (26.0) &  & 13.0 (13.2)\tabularnewline
bathtub &  & 39.8 (32.8) &  & 66.6 (59.1) &  & 73.8 (67.3) &  &  & 24.6 (28.6) &  & 14.1 (17.4)\tabularnewline
bed &  & 36.5 (34.0) &  & 59.3 (56.9) &  & 65.9 (63.8) &  &  & 29.0 (30.5) &  & 16.1 (17.6)\tabularnewline
bench &  & 33.3 (32.4) &  & 56.4 (54.3) &  & 64.0 (61.9) &  &  & 29.4 (30.7) &  & 18.1 (19.2)\tabularnewline
bicycle &  & 26.5 (25.6) &  & 48.5 (47.8) &  & 58.2 (57.9) &  &  & 32.0 (32.2) &  & 23.6 (24.1)\tabularnewline
birdhouse &  & 44.4 (42.1) &  & 71.4 (68.5) &  & 78.0 (75.8) &  &  & 22.2 (23.4) &  & 12.6 (13.3)\tabularnewline
boat &  & 25.7 (22.8) &  & 49.2 (43.7) &  & 58.5 (52.4) &  &  & 32.6 (35.9) &  & 23.1 (27.7)\tabularnewline
bookshelf &  & 40.0 (36.7) &  & 62.7 (59.7) &  & 68.4 (65.6) &  &  & 28.1 (30.0) &  & 14.3 (15.9)\tabularnewline
bottle &  & 57.6 (52.0) &  & 88.1 (84.3) &  & 93.0 (90.5) &  &  & 13.4 (15.0) &  & 10.0 (10.9)\tabularnewline
bowl &  & 42.1 (30.9) &  & 69.4 (55.0) &  & 76.4 (63.6) &  &  & 22.6 (29.5) &  & 13.2 (19.0)\tabularnewline
bus &  & 53.0 (46.4) &  & 79.5 (74.5) &  & 84.4 (80.4) &  &  & 18.2 (21.1) &  & 10.7 (12.0)\tabularnewline
cabinet &  & 55.7 (56.3) &  & 79.5 (79.7) &  & 84.1 (84.1) &  &  & 18.2 (18.2) &  & 10.3 (10.2)\tabularnewline
camera &  & 35.4 (34.9) &  & 61.5 (59.6) &  & 69.7 (67.6) &  &  & 26.4 (27.7) &  & 16.1 (16.7)\tabularnewline
can &  & 64.8 (53.9) &  & 90.6 (81.7) &  & 93.6 (86.6) &  &  & 12.6 (16.9) &  & 9.1 (10.6)\tabularnewline
cap &  & 49.5 (34.1) &  & 79.0 (62.7) &  & 83.9 (70.9) &  &  & 18.9 (26.3) &  & 11.3 (16.0)\tabularnewline
car &  & 33.7 (25.7) &  & 63.1 (51.1) &  & 72.6 (61.2) &  &  & 24.4 (30.7) &  & 16.2 (21.8)\tabularnewline
cellular\_telephone &  & 50.8 (50.3) &  & 78.6 (78.1) &  & 84.8 (84.2) &  &  & 17.8 (18.1) &  & 11.1 (11.2)\tabularnewline
chair &  & 35.1 (33.1) &  & 60.7 (57.7) &  & 68.3 (65.4) &  &  & 27.6 (29.2) &  & 16.4 (17.6)\tabularnewline
clock &  & 41.7 (33.5) &  & 67.3 (57.9) &  & 73.9 (64.9) &  &  & 24.2 (29.6) &  & 13.5 (17.4)\tabularnewline
computer\_keyboard &  & 39.6 (31.5) &  & 71.1 (59.8) &  & 80.0 (69.2) &  &  & 20.5 (26.7) &  & 13.8 (17.5)\tabularnewline
dishwasher &  & 66.9 (64.0) &  & 89.2 (88.2) &  & 91.9 (91.1) &  &  & 13.4 (14.2) &  & 9.0 (9.3)\tabularnewline
display &  & 38.6 (29.3) &  & 67.1 (52.8) &  & 75.1 (60.7) &  &  & 23.3 (31.5) &  & 14.3 (20.4)\tabularnewline
earphone &  & 28.1 (22.4) &  & 53.1 (43.8) &  & 62.8 (53.7) &  &  & 30.1 (35.0) &  & 20.6 (26.9)\tabularnewline
faucet &  & 31.3 (29.5) &  & 58.2 (55.7) &  & 68.2 (65.7) &  &  & 26.8 (28.3) &  & 18.2 (19.3)\tabularnewline
file &  & 59.9 (55.5) &  & 83.1 (81.0) &  & 86.4 (84.9) &  &  & 16.8 (18.1) &  & 9.7 (10.3)\tabularnewline
guitar &  & 38.2 (27.6) &  & 64.6 (48.6) &  & 72.4 (56.6) &  &  & 24.4 (33.5) &  & 15.0 (23.7)\tabularnewline
helmet &  & 36.3 (24.2) &  & 68.6 (50.6) &  & 78.4 (61.3) &  &  & 21.5 (30.7) &  & 14.9 (22.2)\tabularnewline
jar &  & 43.0 (42.7) &  & 71.0 (70.7) &  & 78.0 (77.7) &  &  & 21.6 (21.7) &  & 12.9 (13.0)\tabularnewline
knife &  & 32.9 (28.4) &  & 56.2 (49.8) &  & 65.3 (58.7) &  &  & 28.2 (31.6) &  & 18.6 (22.7)\tabularnewline
lamp &  & 38.3 (36.5) &  & 65.9 (62.6) &  & 73.6 (70.7) &  &  & 24.1 (25.6) &  & 14.6 (15.5)\tabularnewline
laptop &  & 46.2 (31.8) &  & 75.9 (56.3) &  & 82.0 (63.4) &  &  & 20.1 (30.5) &  & 12.1 (18.4)\tabularnewline
loudspeaker &  & 49.7 (47.0) &  & 73.6 (71.9) &  & 79.0 (77.8) &  &  & 20.9 (22.0) &  & 11.3 (12.0)\tabularnewline
mailbox &  & 42.4 (39.4) &  & 68.3 (66.0) &  & 76.0 (74.2) &  &  & 22.5 (23.7) &  & 13.3 (14.3)\tabularnewline
microphone &  & 44.6 (41.3) &  & 74.2 (70.4) &  & 81.4 (78.1) &  &  & 20.1 (22.0) &  & 12.5 (13.5)\tabularnewline
microwave &  & 65.4 (59.4) &  & 87.9 (84.6) &  & 90.8 (88.2) &  &  & 14.2 (16.3) &  & 9.0 (9.8)\tabularnewline
motorcycle &  & 17.9 (16.9) &  & 37.7 (35.8) &  & 48.0 (46.1) &  &  & 37.4 (38.4) &  & 31.7 (33.3)\tabularnewline
mug &  & 54.3 (49.7) &  & 80.8 (78.5) &  & 84.9 (83.5) &  &  & 18.1 (19.2) &  & 10.5 (11.3)\tabularnewline
piano &  & 40.2 (35.7) &  & 65.9 (60.9) &  & 72.8 (68.2) &  &  & 25.2 (27.9) &  & 14.0 (16.1)\tabularnewline
pillow &  & 33.4 (23.4) &  & 66.7 (50.0) &  & 77.3 (60.7) &  &  & 22.3 (31.1) &  & 15.8 (22.5)\tabularnewline
pistol &  & 30.9 (27.2) &  & 57.9 (53.3) &  & 67.7 (63.7) &  &  & 27.1 (29.5) &  & 18.3 (20.6)\tabularnewline
pot &  & 35.5 (33.8) &  & 59.4 (58.2) &  & 66.7 (65.7) &  &  & 28.2 (28.8) &  & 16.5 (17.3)\tabularnewline
printer &  & 46.0 (42.1) &  & 72.1 (68.5) &  & 77.8 (74.5) &  &  & 22.2 (24.4) &  & 12.2 (13.2)\tabularnewline
remote\_control &  & 38.0 (33.3) &  & 65.3 (60.2) &  & 73.8 (69.7) &  &  & 23.9 (26.1) &  & 15.0 (17.3)\tabularnewline
rifle &  & 28.5 (27.1) &  & 53.5 (51.3) &  & 63.3 (61.0) &  &  & 29.5 (30.6) &  & 20.3 (21.7)\tabularnewline
rocket &  & 28.7 (26.6) &  & 54.7 (50.8) &  & 64.7 (60.4) &  &  & 28.6 (31.0) &  & 19.8 (22.0)\tabularnewline
skateboard &  & 45.8 (28.7) &  & 75.9 (54.0) &  & 82.3 (62.1) &  &  & 19.9 (31.0) &  & 12.1 (20.0)\tabularnewline
sofa &  & 37.6 (36.5) &  & 62.2 (60.0) &  & 68.7 (66.3) &  &  & 27.7 (29.2) &  & 15.0 (15.8)\tabularnewline
stove &  & 53.5 (51.4) &  & 78.7 (77.4) &  & 83.9 (82.7) &  &  & 18.3 (19.2) &  & 10.6 (11.0)\tabularnewline
table &  & 53.2 (46.4) &  & 76.7 (72.0) &  & 81.1 (77.1) &  &  & 19.9 (22.8) &  & 10.6 (12.1)\tabularnewline
telephone &  & 50.6 (50.5) &  & 78.7 (78.9) &  & 85.0 (85.4) &  &  & 17.8 (17.7) &  & 11.1 (11.1)\tabularnewline
tower &  & 39.8 (39.2) &  & 65.8 (63.9) &  & 73.0 (71.2) &  &  & 24.5 (25.3) &  & 14.1 (14.5)\tabularnewline
train &  & 34.5 (31.2) &  & 60.1 (56.4) &  & 68.3 (65.0) &  &  & 27.2 (29.1) &  & 16.6 (18.5)\tabularnewline
vessel &  & 26.4 (23.7) &  & 50.0 (45.1) &  & 59.0 (53.7) &  &  & 32.3 (35.3) &  & 22.5 (26.6)\tabularnewline
washer &  & 63.4 (65.0) &  & 86.3 (87.0) &  & 89.7 (90.2) &  &  & 14.9 (14.5) &  & 9.3 (9.1)\tabularnewline
\tabularnewline
\bottomrule
\normalsize
\end{tabular}

\vspace{2mm}
\caption{Surface normal estimation performance for all classes under the $V_d$ view sampling. The numbers outside the parenthesis denote the `Learned' setting and the evaluation in the `Induced' setting is reported in the parenthesis. Higher is better for the first three percent of `good' pixels metrics and lower is better for last two three error metrics.}
\tablelabel{normalResVd}
\end{table*}

\end{document}